\documentclass[lettersize,journal]{IEEEtran}
\usepackage{amsmath,amsfonts}
\usepackage{algorithmic}
\usepackage{algorithm}
\usepackage{array}
\usepackage[caption=false,font=normalsize,labelfont=sf,textfont=sf]{subfig}
\usepackage{textcomp}
\usepackage{stfloats}
\usepackage{url}
\usepackage{verbatim}
\usepackage{graphicx}
\usepackage{cite}

\usepackage{bm}
\usepackage{multirow}
\usepackage{amssymb}
\usepackage{multirow}
\usepackage{slashed}

\usepackage[marginal]{footmisc}

\begin{document}
	
	\title{Improving Vision Anomaly Detection \\ with the Guidance of  Language Modality}
	

	\author{Dong~Chen, Kaihang~Pan, Guoming~Wang, Yueting~Zhuang,~\IEEEmembership{Senior~Member,~IEEE}, Siliang~Tang,    

}

	
	\maketitle
	
	\begin{abstract}
	Recent years have seen a surge of interest in anomaly detection for tackling industrial defect detection, event detection, etc. However, existing unsupervised anomaly detectors, particularly those for the vision modality, face significant challenges due to redundant information and sparse latent space. Conversely, the language modality performs well due to its relatively single data. This paper tackles the aforementioned challenges for vision modality from a multimodal point of view. Specifically, we propose Cross-modal Guidance (CMG), which consists of Cross-modal Entropy Reduction (CMER) and Cross-modal Linear Embedding (CMLE), to tackle the redundant information issue and sparse space issue, respectively. CMER masks parts of the raw image and computes the matching score with the text. Then, CMER discards irrelevant pixels to make the detector focus on critical contents. To learn a more compact latent space for the vision anomaly detector, CMLE learns a correlation structure matrix from the language modality, and then the latent space of vision modality will be learned with the guidance of the matrix. Thereafter, the vision latent space will get semantically similar images closer. Extensive experiments demonstrate the effectiveness of the proposed methods. Particularly, CMG outperforms the baseline that only uses images by $16.81\%$. Ablation experiments further confirm the synergy among the proposed methods, as each component depends on the other to achieve optimal performance. The code for CMG can be found at https://github.com/Anfeather/CMG.
	\end{abstract}
	
	
	\section{Introduction}
	\label{Introduction}

\begin{figure}[t]
	\includegraphics[width=0.47\textwidth]{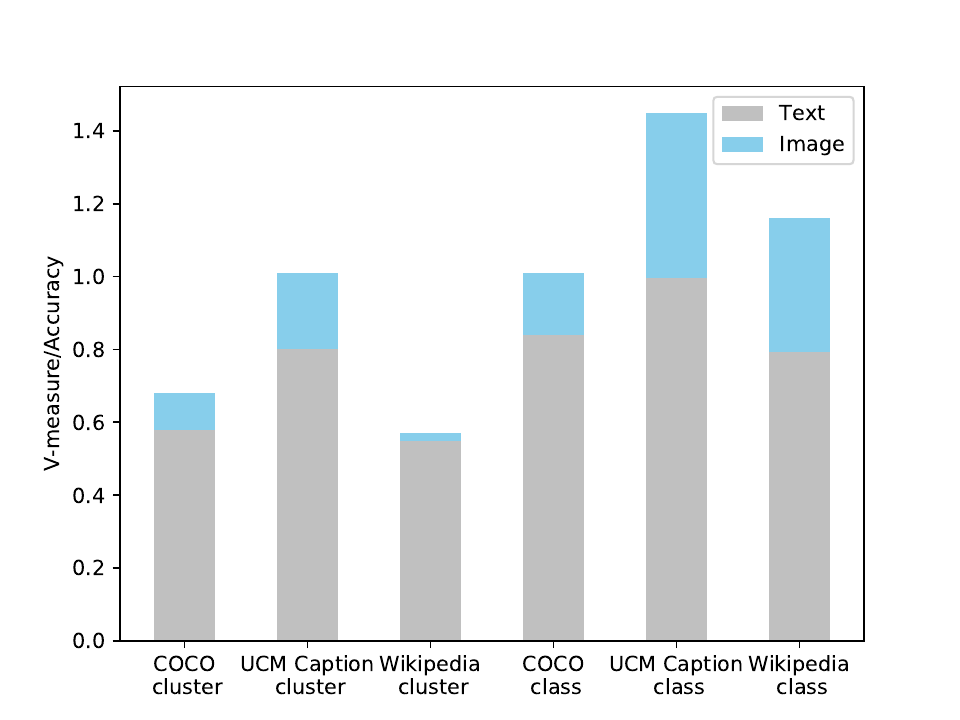}
	\caption{
		The V-measure/Accuracy of different datasets on clustering and classification tasks. ``cluster'' denotes the clustering task, and ``class'' denotes the classification task. The V-measure is the harmonic mean between homogeneity and completeness, which can be used to evaluate the clustering effect \cite{rosenberg2007v}. Images and texts are processed by pre-trained feature extractors, ResNet-50 \cite{he2016deep} and BERT \cite{devlin2018bert}, respectively.}
	\label{data_analysis}
\end{figure}

	\IEEEPARstart{A}{nomaly} detection (AD) is the task of identifying anomalies that differ significantly from the majority of data. The wide applications of AD, such as event detection in earth fields \cite{fisher2017anomaly,wu2018deepdetect}, medical diagnosis and disease detection \cite{schlegl2019f,latif2018phonocardiographic,seebock2019exploiting}, industrial defect detection \cite{bergmann2019mvtec,cui2021sddnet} and video surveillance \cite{zhang2020normality,zhang2022influence,zhou2019attention}, makes it a critical task and have attracted more and more attention. 

One of the most critical research fields of AD is unsupervised AD \cite{zhou2022pull,gao2022robust}, where no prior information on anomalies is available, while there are normal samples for reference. To tackle this problem, three categories of approaches have been proposed: 1) reconstruction-based approaches \cite{hawkins2002outlier,schreyer2017detection}, 2) classification-based approaches \cite{ruff2018deep,geifman2017selective,el2010foundations} and 3) density-based approaches \cite{nalisnick2018deep,kingma2018glow}. 
Additionally, the study \cite{hendrycks2019using} finds that contrastive learning can drastically improve anomaly detection performance on complicated, near-distribution anomalies. Self-supervised Outlier Detection (SSD) \cite{sehwag2021ssd} uses self-supervised representation learning to learn a low dimensional feature space and demonstrates that self-supervised representations are highly effective for AD.

Essentially, all of the aforementioned methods try to learn a discriminative latent space and detect anomalies that are out of distribution in the learned space. However, the latent space of unsupervised samples is usually sparse, especially for visual modalities, due to the considerable variability of samples. For example, images of the same breed of dog may be far apart from each other in the latent space due to factors such as fur color, shooting angle, background, etc., which leads to information sparsity in the latent space. Furthermore, the considerable variability of images makes the correlation structure of samples unavailable, which has been proved to significantly impact the performance of tasks \cite{zhu2017multi}. Such an issue severely degrades the learned boundaries of normal samples and the detector's performance.
Moreover, the vision modality often contains much redundant information, such as the background of surveillance video. The redundant information also degrades the detector's performance, as the learned model may focus more on abundant redundant information. 
In addition, all prior works only focus on a single modality, especially the vision modality, which ignores the numerous multimodal information. 

As depicted in Figure \ref{data_analysis}, there are three multimodal datasets and two tasks. The performance of language modality significantly outperforms that of vision modality in all cases. Specifically, for the clustering task, V-measure shows how well the structural information of the data in the latent space matches the semantic information. The V-measure of language modality is much higher than that of vision modality, suggesting that the latent space of language modality is more informative.
As for the classification task, accuracy reflects the samples' quality of what they describe. In other words, higher-quality (in other words, less redundant information) samples result in higher accuracy. The classification results in Figure \ref{data_analysis} suggest that language is better for describing the target object. Overall, in Figure \ref{data_analysis}, each pair of text and image describes the same object, and the results of texts significantly outperform that of images. Therefore, language modality can help to improve the performance of vision modality when the target tasks are based on structural information and sample quality.

We propose to tackle the aforementioned two challenges in the vision modality with the help of language modality. CMDA \cite{chen2022cross} performs data augmentation with the extra information of multimodal data and improve the performance of the anomaly detector by the augmented data. However, CMDA \cite{chen2022cross} only roughly considers the correlation among samples across different modalities and neglects the fine-grained information and global information provided by multimodal data.

To improve the performance of unsupervised vision AD with the guidance of language modality, we propose Cross-modal Guidance (CMG) that tackles the aforementioned two challenges from two perspectives. 1) Local. Language modality has less redundant information compared to vision modality, and we propose local guidance, Cross-Modal Entropy Reduction (CMER), to reduce redundant information in vision modality. CMER successively masks part of the raw image and calculates matching scores between the remaining content and the text. Then, CMER obtains the best matching masked image with less redundant information. Theoretically, CMER improves the performance of the detector by reducing the entropy of images.
2) Global. Language modality shows a better correlation structure, and we propose global guidance, Cross-modal Linear Embedding (CMLE), where language modality teaches vision modality to construct a compact latent space. Thus, the learned latent space of vision modality will be more compact.

The main contributions of this paper can be summarized as follows:
\begin{itemize}
	\item We discuss redundant information issue and sparse space issue from the multimodal point of view.
	\item With the guidance of language modality, we propose a cross-modal method named CMG, which improves the performance of vision detectors from global and local perspectives.
	\item We extensively evaluate CMG over various datasets, and the proposed method significantly outperforms baselines in most experiments. Specifically, the proposed method significantly outperforms the most important baseline, SSD, by $6.84\%$, $16.81\%$, and $9.21\%$ on Class-COCO, UCM caption, and Wikipedia, respectively.
\end{itemize}

	\section{Related Work}
	Traditionally, anomaly detection can be roughly divided into three themes: classification-based, reconstruction-based, and density-based approaches. Classification-based approaches, such as one-class SVM \cite{li2003improving}, separate the normal samples and the rest of the feature space. Reconstruction-based approaches, such as autoencoder \cite{zhou2017anomaly}, learn the normal distribution by reconstructing input data. Density-based approaches, such as GMM \cite{li2016anomaly}, try to estimate the probability density of samples. All these approaches are well interpretable but can not apply to high-dimensional data. 

For the above problem, some deep anomaly detection methods are proposed, describing the normal training data and scoring anomalies with self-supervision \cite{liznerski2022exposing}. The study \cite{golan2018deep} achieves anomaly detection by augmenting samples with contrastive learning. The self-supervised study \cite{hendrycks2019using} finds that contrastive learning can drastically improve anomaly detection performance on complicated, near-distribution anomalies. Simple contrastive learning (SimCLR) \cite{chen2020simple} creates different augmented views of the same sample with transformations. The augmented views are considered positives, and other samples are considered negatives. CSI \cite{tack2020csi} proposes performing contrastive learning with distributionally-shifted augmentations, where some augmented samples could also be considered negatives. Such augmentations are proven to be beneficial for discriminating normal and anomaly samples. 
Self-supervised Outlier Detection (SSD) \cite{sehwag2021ssd} uses self-supervised representation learning followed by a Mahalanobis distance-based detection in the feature space. SSD demonstrates that self-supervised representations are highly effective for anomaly detection. The proposed framework performs far better than most of the previous unsupervised representation learning methods and performs on par, and sometimes even better, than supervised representations. Although SSD is an excellent framework, it neglects the massive multimodal data. Based on SSD, CMDA \cite{chen2022cross} proposes to perform data augmentation with the extra information of multimodal data, and the proposed method further improves the performance of anomaly detection. Similarly, SHE \cite{zhang2023outofdistribution} also introduces extra information, labels, to detect the anomalies with Hopfield
energy in a store-then-compare paradigm.

Inspired by the effectiveness of CMDA, we propose a new framework, CMG, which improves the learned latent space of vision modality with cross-modal guidance. Specifically, we improve vision anomaly detection by alleviating sparse space and redundant information issues with the guidance of language modality.

	\section{Method}
	\label{sec: method}
	
	This section presents the proposed Cross-modal Guidance (CMG) for vision anomaly detection, which includes local guidance, Cross-modal Entropy Reduction (CMER) and global guidance, Cross-modal Linear Embedding (CMLE). Particularly, CMER and CMLE are designed to reduce redundant information issue and alleviate sparse space issue in the vision modality with the guidance of the language modality. 
	
	\subsection{Problem Formulation}
	\label{setup}
	The studied problem, unsupervised vision anomaly detection, can be formally stated as follows. Given training normal images, \textbf{X}, corresponding texts, \textbf{Y}, single image, $x_i\in$ \textbf{X}, and single text, $y_i\in$ \textbf{Y}. Conventional vision anomaly detection aims to train a detector with \textbf{X} and distinguish abnormal images that deviates from the learned distributions. Differently, CMG learns vision anomaly detector from global and local perspective with \textbf{Y} and $y_i$, respectively. During inference, CMG distinguishes anomalies solely based on the vision modality.
	
	\subsection{Local Guidance: Cross-modal Entropy Reduction}
	\begin{figure}[h]
		\includegraphics[width=0.47\textwidth]{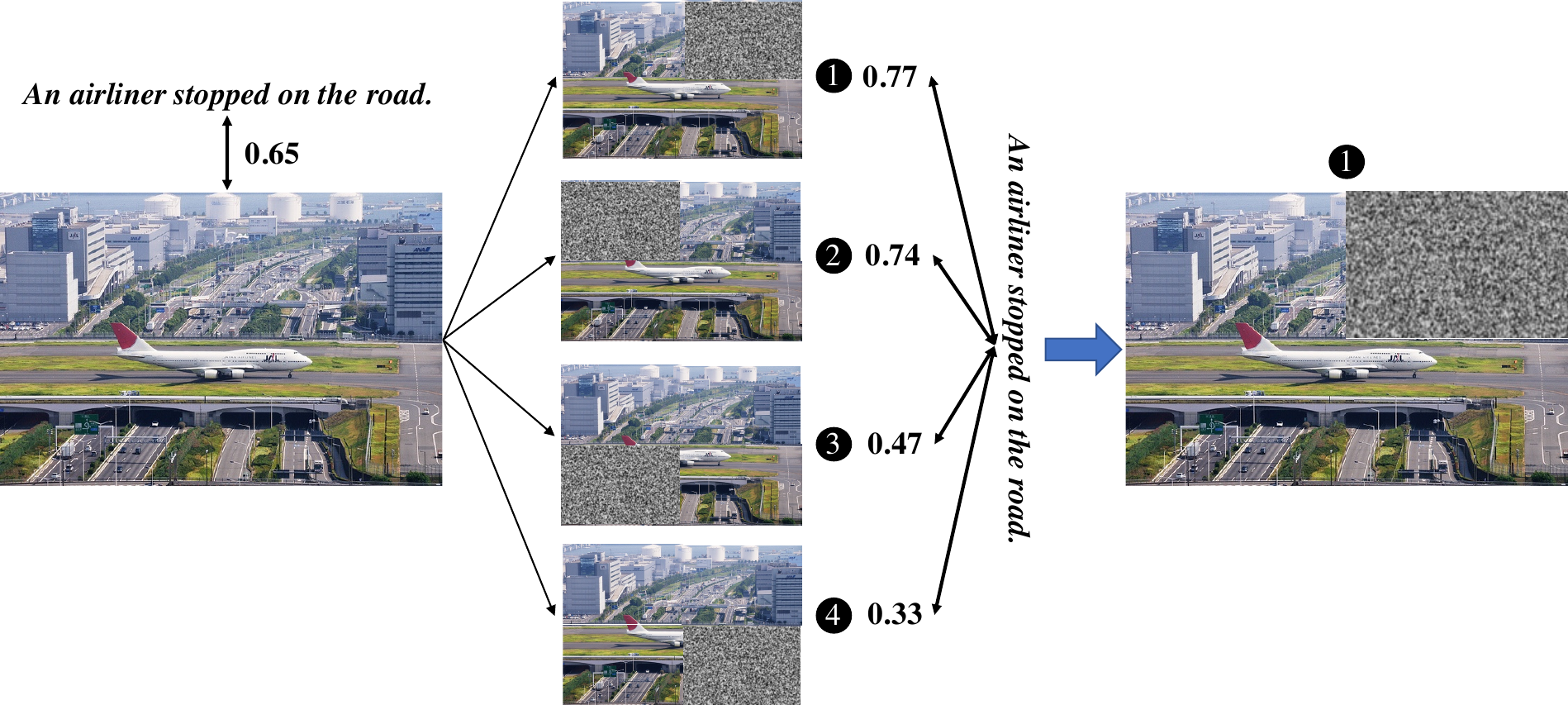} 
		\caption{Cross-modal Entropy Reduction (CMER). CMER masks parts of the raw image and compares the remaining contents with text to select the most matching masked image. In this instance, the raw image is divided into four parts and different parts are masked in turn.}
		\label{LG}
	\end{figure}
	This section addresses the issue of redundant information from a local perspective.
	
	Based on the previous analysis, images contain much more redundant information, such as background or redundant objects, which severely impacts the performance of the anomaly detector, as it cannot focus on important content. Figure \ref{LG} illustrates the proposed CMER, which reduces the redundant information by successively masking parts of the raw image and computing the matching score with the paired text. In this way, we can get the more important parts of the raw image.
	
	In order to mask redundant information for each image, we focus on single image, $x_i$ and the corresponding text, $y_i$.
	To compute the matching score between images and texts, we first train the feature extractor, $E_1^I$, for images, which projects all samples into the latent space of the pretrained text model, $E_1^T$, by contrastive learning \cite{chuang2020debiased}:
	
	\begin{equation}
		\begin{aligned}
			&\ell_{i} =-\log \frac{ \exp \left(\operatorname{sim}\left(\boldsymbol{z}_{i}^{T}, \boldsymbol{z}_{i}^{I}\right) / \tau\right)}{\sum_{j=1}^{m}  1_{[j \neq i]} \exp \left(\operatorname{sim}\left(\boldsymbol{z}_{i}^{T}, \boldsymbol{z}_{j}^{I}\right) / \tau\right)} \\
			\label{CL}
		\end{aligned}
	\end{equation}
	where $x_i$ and $y_i$ are a pair, $\boldsymbol{z}_{i}^{I} = E_1^I(x_i)$, $\boldsymbol{z}_{i}^{T} = E_1^T(y_i)$, $sim(\cdot,\cdot)$ is cosine similarity, $1_{[j \neq i]} \in\{0,1\}$ is an indicator evaluating to 1 iff $j \neq i$, $\tau$ denotes a temperature parameter and $m$ is the number of samples. Eq.(\ref{CL}) optimize $E_1^I$ and $E_1^T$ by pulling each pair of the same sample close while pushing away from other samples.
	
	After training with Eq.(\ref{CL}), the matching score between text $i$ and image $j$ is computed as:
	
	\begin{equation}
		\begin{aligned}
			&s_{ij} = \frac{ \operatorname{sim}\left(\boldsymbol{z}_{i}^{T}, \boldsymbol{z}_{j}^{I}\right) }{\sum_{i=1}^{m} \sum_{j=1}^{m}  \operatorname{sim}\left(\boldsymbol{z}_{i}^{T}, \boldsymbol{z}_{j}^{I}\right) } \\
			\label{score}
		\end{aligned}
	\end{equation}
	
	Thereafter, for a raw image $x$, we mask its pixels with two strategies: hard mask and soft mask. Specifically, the hard mask discards all masked region information by setting it to 0. In contrast, the soft mask keeps some raw pixel information by timing a small constant. Next, there will be $M$ masked samples, i.e., $[\overline{x_1},\overline{x_2}, ... ,\overline{x_M}]$. For example, in Figure \ref{LG}, we divide the raw image into four parts, mask one of them in turn, and compute the text-image matching score by Eq.(\ref{score}). Then, the masked image with the highest matching score is selected for anomaly detection, and the subscript is labeled $l$. For Figure \ref{LG}, $l=[1,0,0,0]$.
	
	With $\overline{x_l}$, we can learn a compact and informative latent space, as images have been stripped of redundant information, and the distribution in the latent space relies more on meaningful information. However, such a process is not available when testing due to the absence of the language modality. Therefore, we propose to train a Redundant Information Detector (RID) for CMER to predict the redundant region without the language modality. During training of RID, masked samples and labels $l$ are input into RID, and the cost function is:
	
	\begin{equation}
		\mathcal{L}_{RID} = \sum_{i=1}^{M}l_i\frac{\exp(F_{RID}(\overline{x_i}))}{\sum_{m=1}^{M}\exp(F_{RID}(\overline{x_m}))}
		\label{RID}
	\end{equation}
	
	Note that, Eq. (\ref{RID}) is solely used for training the RID model and does not participate in the learning of the latent space.
	During testing of detection, we use RID to predict the masked region and get the masked image $\widetilde{x}$.

	We further perform theoretical analysis from an entropy perspective to show that CMER can effectively reduce redundant information. We first present a lemma:
	\begin{equation}
		\begin{aligned}
			&\sum_{i=1}^n a_i \log _2 \frac{a_i}{b_i}-\operatorname{alog}_2 \frac{a}{b} \\
			&=a\left[\sum_{i=1}^n \frac{a_i}{a} \log _2 \frac{a_i}{b_i}-\left(\sum_{i=1}^n \frac{a_i}{a}\right) \log _2 \frac{a}{b}\right] \\
			&=a\left[\sum_{i=1}^n \frac{a_i}{a} \log _2 \frac{a_i}{b_i}-\sum_{i=1}^n \frac{a_i}{a} \log _2 \frac{a}{b}\right] \\
			&=a\left[\sum_{i=1}^n \frac{a_i}{a} \log _2\left(\frac{a_i}{b_i} \cdot \frac{b}{a}\right)\right] \\
			&\geq a\left[\sum_{i=1}^n \frac{a_i}{a} \log _2(e)\left(1-\frac{b_i}{a_i} \cdot \frac{a}{b}\right)\right] 
		\end{aligned}
	\end{equation}
	where $a$, $b$ are samples from different modalities, $n$ is the number of samples in one modality.
	Therefore,
	\begin{equation}
		\begin{aligned}
			&a=\sum_{i=1}^n a_i, \quad b=\sum_{i=1}^n b_i \\
			&\sum_{i=1}^n a_i \log _2 \frac{a_i}{b_i} \geq a \log _2 \frac{a}{b}
		\end{aligned}
	\end{equation}

	Let $\mathbb{X}$ and $\mathbb{Y}$ be the variable of image and text, respectively. We use entropy to represent the content of redundant information in samples; the higher the entropy, the more redundant information. $H(\mathbb{X})$ is the entropy of the image and $H(\mathbb{Y})$ is the entropy of the text. Additionally, we use $H(\mathbb{X}\mid \mathbb{Y})$ to represent the entropy of the masked image, i.e., $\overline{\mathbb{X}} =\mathbb{X}\mid \mathbb{Y}$, as the masked image is built on image $\mathbb{X}$ with the condition text $\mathbb{Y}$. 
	Now we show that with text $\mathbb{Y}$, the entropy of the masked image $H(\overline{\mathbb{X}})$ is lower than that of the raw image $H(\mathbb{X})$:
	
	\begin{equation}
		\begin{aligned}
			& H(\mathbb{X})-H(\overline{\mathbb{X}}) \\
			&=\sum_{}-p(x_i) \log _2 p(x_i)-\sum_{} \sum_{}-p(x_i, y_i) \log _2 p(x_i \mid y_i) \\
			&=\sum_{} \sum_{} p(x_i, y_i) \log _2 \frac{p(x_i, y_i)}{p(x_i) p(y_i)}
		\end{aligned}
		\label{entropy}
	\end{equation}
	According to the lemma, Eq.(\ref{entropy}) can be further written as:
	
	\begin{equation}
		\begin{aligned}
			&\sum_{} \sum_{} p(x_i, y_i) \log _2 \frac{p(x_i, y_i)}{p(x_i) p(y_i)}\\
			&\geq\left[\sum_{} \sum_{} p(x_i, y_i)\right] \log _2 \frac{\sum_{} \sum_{} p(x_i, y_i)}{\sum_{} \sum_{} p(x_i) p(y_i)}=0
		\end{aligned}
	\end{equation}
	Note that the above inequality takes the equal sign iff $\mathbb{X}$ and $\mathbb{Y}$ are independent, i.e., $p(x,y)=p(x)p(y)$. However, in this work, image $x$ and text $y$ are highly correlated as they are pairs. Thus, we get $H(\mathbb{X}) > H(\overline{\mathbb{X}})$, which validates that the masked image in CMER gets lower entropy by reducing redundant information.
	
	\subsection{Global Guidance: Cross-modal Linear Embedding}
	\begin{figure}[h]
		\includegraphics[width=0.47\textwidth]{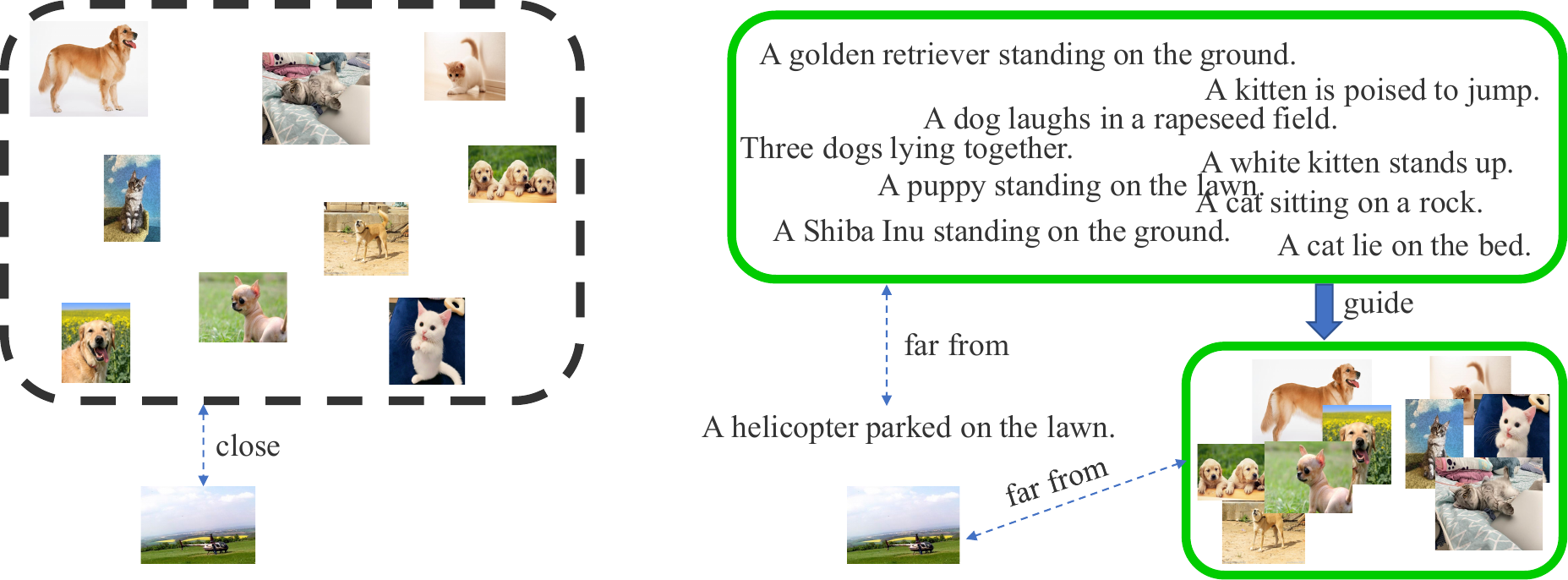}
		\caption{Cross-modal Linear Embedding (CMLE). Unsupervised methods usually get visually similar but semantically dissimilar images closer in the latent space, where the correlation structure is discarded, and the samples are sparse. CMLE improves the latent space of images by providing the correlation structure of texts.}
		\label{GG}
	\end{figure}
	Data usually contain the correlation structure, which is helpful for computer vision tasks \cite{zhu2017multi}.
	In this section, we further consider the correlation structure of the language modality, which is helpful in making visually and semantically similar images more similar and visually similar but semantically dissimilar images more dissimilar. 
	Figure \ref{GG} illustrates the main idea. Left of Figure \ref{GG} shows the latent space without guidance of language modality. Intuitively,  even if the content is completely unrelated, images with similar backgrounds may have closer distances in the latent space. For example, in the left image, a helicopter parked on a lawn may be closer to a dog on the lawn. In contrast, with the guidance of language modality, the distribution of images in latent space will adhere more closely to the main content.  In the right of Figure \ref{GG}, image of helicopter will maintain a significant distance in the latent space.
	
We use boldface uppercase letters to denote all samples and lowercase letters to denote a single sample, while $y_i$ denotes the representation of the ith text in the latent space instead of raw text in this section. 
	Similar to Locally Linear Embedding \cite{roweis2000nonlinear}, we first try to find similar samples, for which we perform clustering for texts. Then, there are $K$ groups, \textbf{Y} = $[\mathbf{Y_1}, \mathbf{Y_2}, ..., \mathbf{Y_K}]$. For one group, $ \mathbf{Y_k}$, we represent the correlation structure among texts with a matrix, \textbf{W}, that is,
	\begin{equation}
		\begin{aligned}
			y_i = \sum_{j=1}^{\mathcal{N}\left(Y_k\right)  }\mathbf{W}_{ij} y_{j}
		\end{aligned}
	\end{equation}
	where, $\mathcal{N}\left(Y_k\right)$ is the number of samples belongs to group $k$. 
	
	Then, the correlation structure matrix $\mathbf{W}$ of the $k$th group can be calculated as follows:
	\begin{equation}
		\begin{aligned}
			&\epsilon(\mathbf{W})=\sum_{i=1}^{\mathcal{N}\left(Y_k\right)}\left\|y_i-\sum_{j\neq i}^{\mathcal{N}\left(Y_k\right)  }\mathbf{W}_{ij} y_{j}\right\|_2^2,
			\text { s.t. } \mathbf{W_{i,*}^\top}\mathbf{1} =1 
		\end{aligned}
	\end{equation}
	where $\mathbf{W_{i,*}}$ denotes the $i$th row of $\mathbf{W}$. For convenience, let $w_i=\mathbf{W_{i,*}} \in \mathbb{R}^{ \mathcal{N}\left(Y_k\right) \times 1}$, and for a single sample of $k$th group, $y_i$, our objective becomes:
	
	\begin{equation}
		\begin{aligned}
			&w_i=\underset{w_i}{\arg \min } \left\|y_i-w_i^\top Y_k\right\|^2_2,
			\text { s.t. } w_i^\top\mathbf{1} =1
		\end{aligned}
	\end{equation}
	Use Lagrange multiplier method as:
	\begin{equation}
		\begin{aligned}
			\mathcal{L}&=\sum_{i=1}^{\mathcal{N}\left(Y_k\right) }w_i^{\top} G_i w_i-\sum_{i=1}^{\mathcal{N}\left(Y_k\right) } \lambda_i\left(w_i^\top\mathbf{1}-1\right) 
			\label{L}
		\end{aligned}
	\end{equation}
	then partially differentiating $\mathcal{L}$ in Eq.(\ref{L}) with respect $w_i$ and $\lambda_i$, after simplification, we can get the correlation structure matrix by:
	\begin{equation}
		w_i=\frac{G_i^{-1}\mathbf{1}}{\mathbf{1}^\top G_i^{-1}\mathbf{1}}
		\label{correlation S}
	\end{equation}
	
	With correlation structure matrix computed by Eq.(\ref{correlation S}), we can guide feature extractor of images, $E_{2}^I$, with the correlation structure matrix, $\mathbf{w}$:
	\begin{equation}
		\begin{aligned}
			\mathcal{L}_{global} = \|E_{2}^I(x)-w E_{2}^I(x)\|_2^2\\
			\label{Loss_GG}
		\end{aligned}
	\end{equation}
	\subsection{Cross-modal Guidance}
		\begin{figure*}[t]
		\centering
		\includegraphics[width=0.65\textwidth]{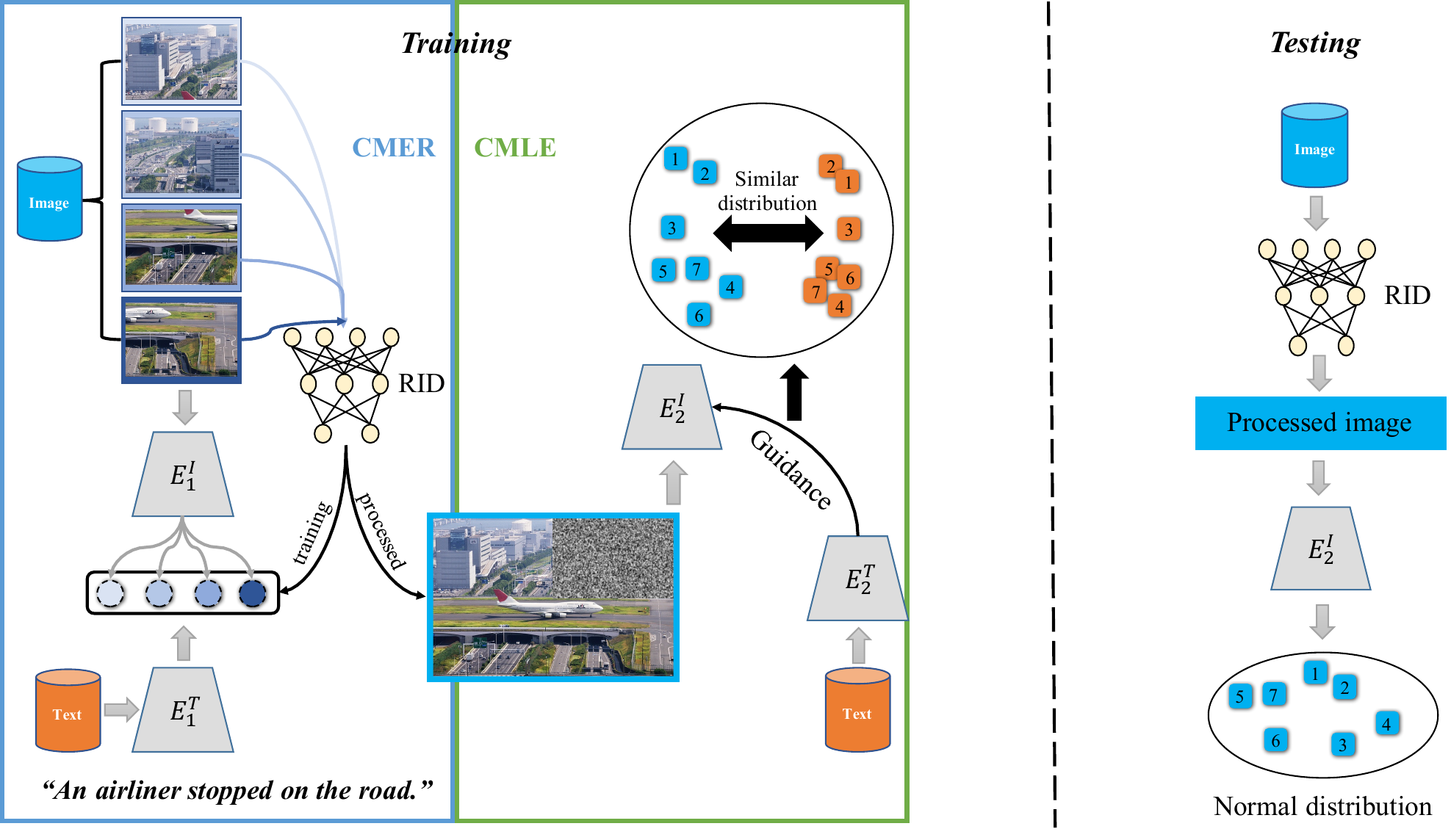}
		\caption{The whole architecture of Cross-modal Guidance (CMG).}
		\label{CMG_whole}
	\end{figure*}
	
	The whole architecture of CMG is illustrated in Figure \ref{CMG_whole}. During training, we first train the cross-modal matching model, $E_1^I$. Then, we mask raw images with the scale value and compute the matching scores among masked images. With masked images and matching scores, we can train the RID model, $F_{RID}$, to predict the masked region during testing. Next, we train the image extractor, $E_2^I$, with the guidance of language modality to learn a compact and informative latent space. In addition, we use the Mahalanobis distance to calculate the anomaly score in the latent space of $E_2^I$:
	
	\begin{equation}
		s_a = \underset{k}{\arg \min }(E_2^I(\overline{x})-\mu_k)^T\Sigma_{k}^{-1}(E_2^I(\overline{x})-\mu_k)
		\label{anomaly score}
	\end{equation}
	where $\mu_k$ and $\Sigma_k$ are the sample mean and sample covariance of features of the normal training samples.
	
	Intuitively, CMER and CMLE are in a mutually cooperative relationship, where CMER provides $E_2^I$ with images that have less redundant information, allowing $E_2^I$ to better focus on the structural relationships provided by CMLE. Meanwhile, CMLE prevents $E_2^I$ from excessively focusing on masked information from CMER (i.e., discriminating normal samples or anomalies  relying on masked regions). Therefore, through the joint action of CMER and CMLE, $E_2^I$ can learn a more compact image latent space. The training and testing process is summarized in Algorithm \ref{CMG_A} and \ref{CMG_B}, respectively.
	
	\begin{algorithm}[!h]
		\caption{The training of CMG }
		\begin{algorithmic}[1]
			\STATE For a given dataset with image modality data $X$ and language modality data $Y$. 
			\STATE Train cross-modal matching models $E_1^I$ and $E_1^T$ by Eq.(\ref{CL}).
			\STATE Mask parts of raw image, compute the matching score between masked image $\overline{x}$ and text $y$ with Eq.(\ref{score}) and record the subscript of masked region as label $l$.
			\STATE Train RID with $l$, $\overline{x}$ by Eq.(\ref{RID}).
			\STATE Perform clustering for $Y$ and get $K$ groups. 
			\FOR{$k$ in $K$} 
			\STATE	Learn the correlation structure, $w$, by Eq.(\ref{correlation S}) .
			\ENDFOR
			\STATE Learn a latent space with $\overline{x}$ and $w$ for images by Eq.(\ref{Loss_GG}).
		\end{algorithmic}
		\label{CMG_A}
	\end{algorithm}

	\begin{algorithm}[!h]
		\caption{The testing of CMG }
		\begin{algorithmic}[1]
			\STATE For a given dataset with image modality data $X$ and language modality data $Y$. 
			\STATE  Predict masked region and get the masked image $\widetilde{x}$ with RID.
			\STATE Project $\widetilde{x}$ into the latent space with $E_2^I$ .
			\STATE Discriminate anomalies and normal samples with Eq.(\ref{anomaly score}).
		\end{algorithmic}
		\label{CMG_B}
	\end{algorithm}

	\section{Experiments}
	In our experiments, we aim to 
	1) validate the effectiveness of CMG on different datasets, 
	2) validate that CMG can reduce redundant information, 
	3) validate that CMG can learn a compact latent space, 
	4) validate that there is a cooperative relationship between CMER and CMLE.
	
	In our experiments, we mainly compare three types of recent methods to show the introduced texts and the proposed method is effective: 
	a) no extra information except the images, SSD \cite{sehwag2021ssd},
	b) with texts besides images, CMDA \cite{chen2022cross},
	c) with class labels besides images, SHE \cite{	zhang2023outofdistribution}.
	By the way, the proposed CMG belongs to b).
	
	\subsection{Datasets and Settings}
	In this work, we conduct experiments on three different multimodal datasets that with clear class distinctions to divide normal samples and anomalies.
	
	Class-COCO. This dataset is proposed in \cite{chen2022cross}, which is based on the MS COCO \cite{lin2014microsoft}. We follow \cite{chen2022cross} to select six groups, with 45,205 normal samples, while 15,150 samples of 50 groups are regarded as anomalies.
	
	UCM caption. This data set is proposed in \cite{qu2016deep}, which contains 21 classes land use images, and 5 different sentences are exploited to describe every image. We randomly select 6 classes to be normal samples, while 15 classes to be anomalies.
	
	Wikipedia\cite{rasiwasia2010new}. There are 2,866 image-text pairs that belong to 10 classes. We divide the dataset into normal samples (4 classes) and abnormal samples (6 classes). 
	
	\subsection{ Implementation Details}
	We train the ResNet-50 \cite{he2016deep} as feature extractor for the vision anomaly detector, and all texts are projected into the latent space by a pretrained BERT \cite{xiao2018bertservice}. We also train a lightweight network that includes two fully-connected layers to project text vectors into the common latent space for texts and images. Each layer of fully-connected layers follows a ReLU layer except the last one, with 2,048 and 128 hidden units, respectively. 
	As for RID, we use a lightweight network with three fully-connected layers, containing 512, 256, 128 hidden units, to predict the useless region. For an input image, we sequentially mask different regions and input them to RID. RID will select the masked image that with the least redundant information.
	For Class-COCO, we follow \cite{chen2022cross} to employ stochastic gradient descent as the optimizer with a learning rate of 0.01 for 200 epochs, weight decay of 1e-4, and a batch size of 128. Moreover, for Wikipedia and UCM caption, we employ Adam \cite{kingma2014adam} with a learning rate of 0.0001, $\beta_1=0.5$, and $\beta_2=0.9$. The number of parts to mask is 4 for Class-COCO and UCM, and 9 for Wikipedia. The number of mixture components to perform clustering is 5. 
	Note that, for Wikipedia, we conduct experiments on raw images instead of processed data, thus the results of CMDA are different from that of \cite{chen2022cross}.

	\subsection{Results of CMG}
	We show the results of CMG and baselines in Table \ref{Table:main}. Compared to established baselines that ignore the global or local information of multimodal data, our approach yields significant performance improvements in most cases. Although CMDA outperforms CMG on Class-COCO, CMG surpasses CMDA on UCM caption and Wikipedia. Such results are caused by the differences in different datasets. For Class-COCO, there are more objects in images and captions, which benefits CMDA for more meaningful data augmentation. However, the images and captions are similar in the UCM caption, which makes the interpolated images meaningless. As for Wikipedia, the images are too broad, degrading the argument data and the performance of CMDA (only a slight improvement compared to SSD). In contrast, CMG does not have too many requirements for the datasets. It significantly outperforms the most important baseline, SSD, by $6.84\%$, $16.81\%$, and $9.21\%$ on Class-COCO, UCM caption, and Wikipedia, respectively. 
	In addition, the performance of SHE is relatively poor or even unable to work on the dataset, Wikipedia. SHE detects the OOD sample with Hopfield energy in a store-then-compare paradigm, where patterns are stored to represent classes. However, in our experiments, the datasets are more complex and cannot be classified well. Thus, the stored patterns are vague and cannot measure the discrepancy of unseen data. Such results indicate that CMG does not rely on the text's class information, but on structural information to improve the latent space for the vision modality.
	\begin{table}[h]
		\centering
		\caption{Results of the proposed method and baselines.  The $\pm$ shows $95\%$ confidence interval over tasks. }
		\begin{tabular}{ccccccccc}
			\hline
			\multirow{2}{*}{Dataset}& \multirow{2}{*}{Method} & \multirow{2}{*}{AUROC}& \multirow{2}{*}{AUPR} \\
			\\
			\hline
			\multirow{4}{*}{Class-COCO} 
			&SSD	& 		$76.46 \pm 1.16\%$ & $88.59 \pm 0.69\%$ \\
			&CMDA & $\bm{83.25 \pm 1.25\%}$ &${91.96 \pm 1.06\%}$  \\
			&SHE&	$72.15\%$ & $\bm{92.80\%}$\\
			&CMG & ${81.69 \pm 0.99\%}$ &${91.39 \pm 0.79\%}$  \\
			\hline
			\multirow{4}{*}{UCM caption} 
			&SSD	& 		$85.36 \pm 1.36\%$ & $98.20 \pm 0.20\%$ \\
			&CMDA & ${93.37 \pm 1.47\%}$ &${99.28 \pm 0.18\%}$  \\
			&SHE&	$79.97\%$ & $97.67\%$\\
			&CMG & $\bm{99.71 \pm 0.21\%}$ & $\bm{99.98 \pm 0.01\%}$ \\
			\hline
			\multirow{4}{*}{Wikipedia} 
			&SSD	& 		$60.06 \pm 1.46\%$ & $88.22 \pm 0.32\%$ \\
			&CMDA & ${61.29 \pm 4.89\%}$ &${88.76 \pm 2.26\%}$  \\
			&SHE&	$51.57\%$ & $87.41\%$\\
			&CMG & $\bm{65.59 \pm 1.19\%}$ &$\bm{89.82 \pm 0.82\%}$  \\
			\hline
		\end{tabular}
		\label{Table:main}
	\end{table}

	\subsection{CMG Reduces Redundant Information}
	
	In this section, we empirically validate that CMG can reduce redundant information in raw images.
	
	Assume that the redundant information in images is noise and follows a Gaussian distribution \cite{chen2022robust}. We roughly quantify the redundant information through distance correlation, which measures the dependence between two paired random vectors of arbitrary, not necessarily equal, dimensions. The larger the value of distance correlation, the higher the correlation between the paired vectors. Therefore, if the distance correlation between masked images and noise is lower than that of raw images and noise, then the masked images contain less redundant information.
	\begin{figure}[h]
		\includegraphics[width=0.47\textwidth]{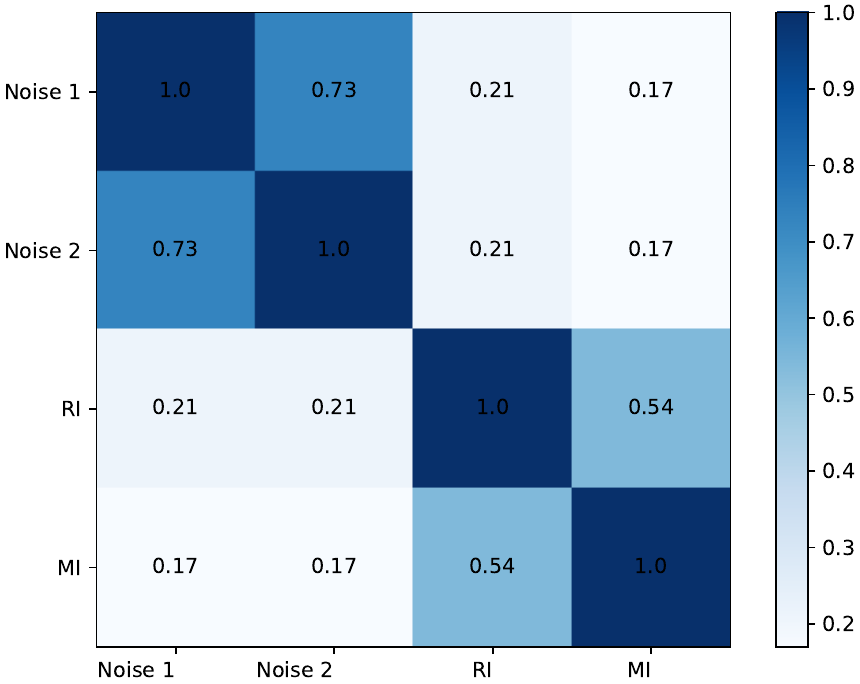}	
		\caption{Heatmap of distance correlation among masked image, raw image and noise. Noise 1 denotes white noise, Noise 2 denots uniform distribution in [0,1], RI denotes the raw image, MI denotes the image masked by CMER.}	
		\label{Figure:heatmap}
	\end{figure}	
	
	As illustrated in Figure \ref{Figure:heatmap}, the distance correlation between raw images and noise is $0.21$, while the distance correlation between masked images and noise is $0.17$. It implies that raw images are more related to noise, and CMG can effectively reduce redundant information in raw images.
	
	\subsection{CMG Learns More Compact Latent Space}
\begin{figure}[h]
		\includegraphics[width=0.47\textwidth]{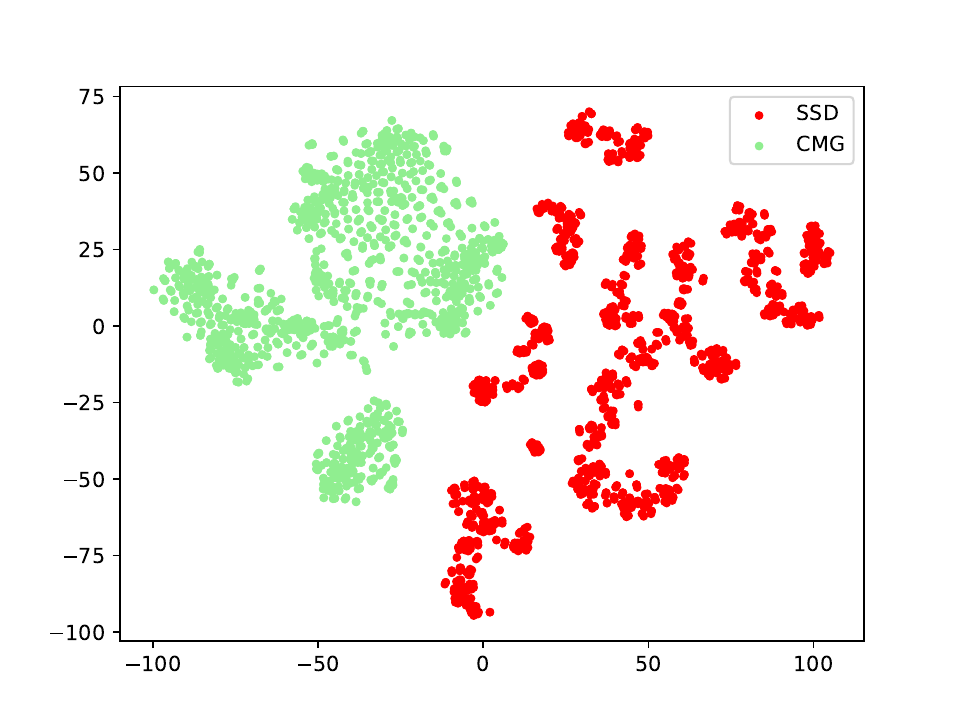}	
		\caption{The latent space learned by CMG and SSD. The dimensionality-reduction algorithm for visualizing high-dimensional data sets is t-SNE \cite{van2008visualizing}.}	
		\label{Figure:t-SNE}
		
	\end{figure}

	To validate that CMG can learn a compact latent space (i.e., alleviate the sparse space issue), we visualize the latent space learned by CMG and SSD in Figure \ref{Figure:t-SNE}. Obviously, the latent space learned by CMG (the green data points) is more compact, better preserving the clusters' global alignment. Whereas the latent space learned by SSD (the red data points) is messier, as contrastive learning focuses on visually similar but ignores semantically similar. Therefore, with language modality guidance, CMG can effectively alleviate the sparse space issue.
	
	\subsection{Ablation}
	\label{sec:ablation}
	We show the effectiveness of CMER and CMLE in Table \ref{Table:ablation}. It can be seen that CMG (SSD+CMER+CMLE) significantly outperforms SSD on all datasets, particularly by $16.81\%$ on AUROC of UCM caption. Besides, SSD+CMLE also surpasses SSD on all datasets, but the improvement is relatively small. However, SSD+CMER even degrades the performance of SSD on Class-COCO and Wikipedia. This result may be caused by the same masked region in CMER. More specifically, CMER masks parts of raw images with scale values (such as 0), and the same mask region can make semantically dissimilar images more visually similar. For instance, during training, model $E_2^I$ pulls images that have the upper left corner masked close in the latent space, as all values of their upper left corner are 0. During testing, $E_2^I$ may project an anomaly to the same position when its upper left corner is also masked. In contrast, when combining CMER with CMLE, $E_2^I$ learns the latent space with a correlation structure matrix, which helps the model ignore the position of masked region. The significant improvement of CMER+CMLE compared to the single method validates that there is a cooperative relationship between CMER and CMLE.

	\begin{table}[t]
		\centering
		\caption{Results of ablation experiments. The $\pm$ shows $95\%$ confidence interval over tasks. }
		\label{Table:ablation}
		\begin{tabular}{ccccccccc}
			\hline
			\multirow{2}{*}{Dataset}& \multirow{2}{*}{Method} & \multirow{2}{*}{AUROC}& \multirow{2}{*}{AUPR} \\
			\\
			\hline
			\multirow{4}{*}{Class-COCO} 
			&SSD& 		$76.46 \pm 1.16\%$ & $88.59 \pm 0.69\%$ \\
			&SSD+ER&	$75.35 \pm 1.45\%$ & $87.97 \pm 0.87\%$\\
			&SSD+LE& ${80.41 \pm 0.91\%}$ & ${90.24 \pm 1.24\%}$ \\
			&CMG& $\bm{81.69 \pm 0.99\%}$ & $\bm{91.39 \pm 0.79\%}$ \\
			\hline
			\multirow{4}{*}{UCM caption} 
			&SSD& 		$85.36 \pm 1.36\%$ & $98.20 \pm 0.20\%$ \\
			&SSD+ER&	$95.45 \pm 1.51\%$ & $99.53 \pm 0.14\%$\\
			&SSD+LE& ${93.44 \pm 1.14\%}$ & ${99.32 \pm 0.12\%}$ \\
			&CMG& $\bm{99.71 \pm 0.21\%}$ & $\bm{99.98 \pm 0.01\%}$ \\
			\hline
			\multirow{4}{*}{Wikipedia} 
			&SSD& 		$60.06 \pm 1.46\%$ & $88.22 \pm 0.32\%$ \\
			&SSD+ER&	$55.08 \pm 4.78\%$ & $86.64 \pm 1.04\%$\\
			&SSD+LE& ${61.97 \pm 1.17\%}$ & ${89.40 \pm 0.70\%}$ \\
			&CMG& $\bm{65.59 \pm 1.19\%}$ & $\bm{89.82 \pm 0.82\%}$ \\
			\hline
		\end{tabular}
	\end{table}

\begin{figure}[h]
		\includegraphics[width=0.47\textwidth]{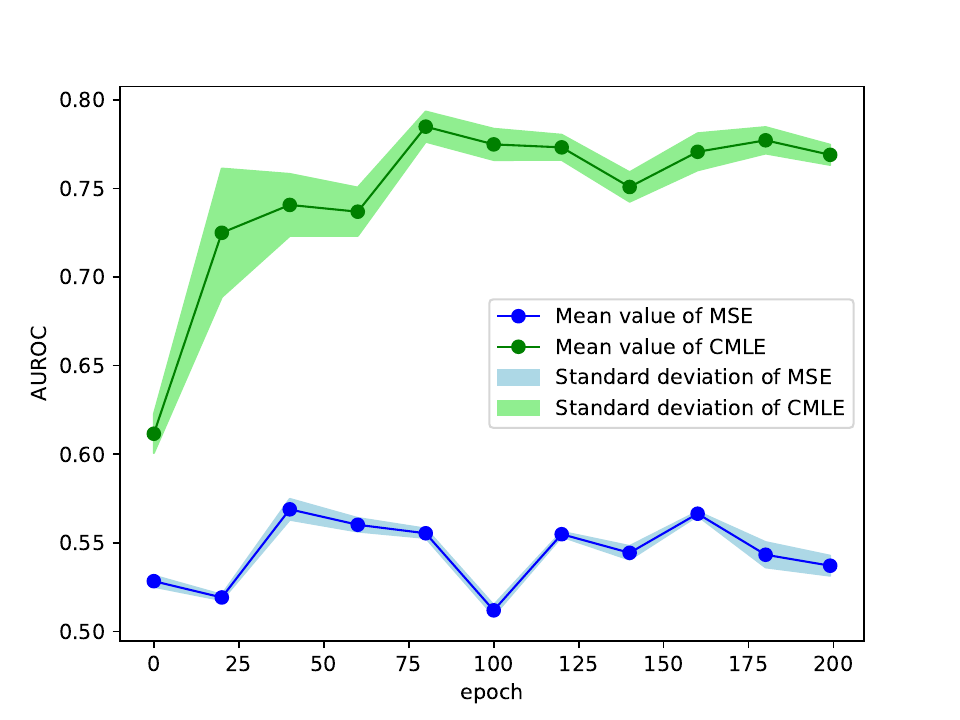}	
		\caption{Compare CMLE with MSE.}	
		\label{Figure:compare_with_MSE}
	\end{figure}

	We also compare the correlation structure matrix in CMLE with a more straightforward method, i.e., distillation with Mean Square Error (MSE). For the latter method, we make the features of images and the features of texts as close as possible with MSE. The results are illustrated in Figure \ref{Figure:compare_with_MSE}, where the standard deviation of MSE is small and the AUROC is lower than that of CMLE. This indicates that guidance with MSE is more stable but cannot provide practical information of language modality for vision modality. The ineffectiveness of MSE may result from the vast discrepancy between language and vision modality, which leads to overfitting. Thus, the proposed CMLE is more effective by guiding vision modality with correlation structure of language modality.
	
	\subsection{Analysis of RID}

\begin{figure}[h]
		\includegraphics[width=0.47\textwidth]{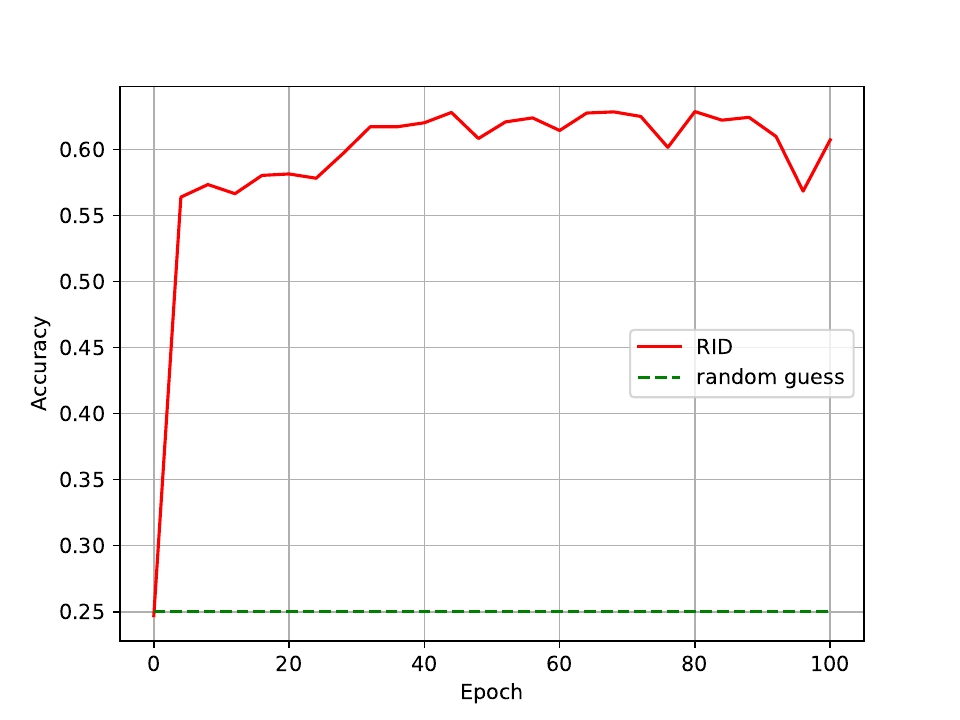}	
		\caption{The convergence of RID.}	
		\label{Figure:convergence}
	\end{figure}

	In this section, we first present the convergence of RID by Figure \ref{Figure:convergence} to show that the idea of predicting masked region is working. 
	
	For Class-COCO, there are four regions that can be masked. Thus, it is a 4-way classification task. It can be seen that the performance of RID rapidly rises with the epoch increasing and eventually reaches around 0.6. The accuracy of RID, 0.6, significantly surpasses random guessing, 0.25, which validates the effectiveness of RID.
	Note that parts of the incorrect prediction of RID will not degrade the performance of the proposed method. As shown in Figure \ref{LG}, whatever masked image 1 or 2, most effective information is retained, and much redundant information is removed. RID tries to reduce redundant information as much as possible. The accuracy of RID, $60\%$, is the probability that it masks the region containing the most redundant information (i.e., masked image 1 in \ref{LG}). However, if RID generates masked image 2 (i.e., incorrect prediction), the masked image is undoubtedly still helpful to the proposed algorithm. In contrast, the probability that RID masks the region containing the most effective information is much less than $40\%$, which can be inferred from the excellent results of CMG in Table 1.
	
	\begin{figure}[h]
		\includegraphics[width=0.47\textwidth]{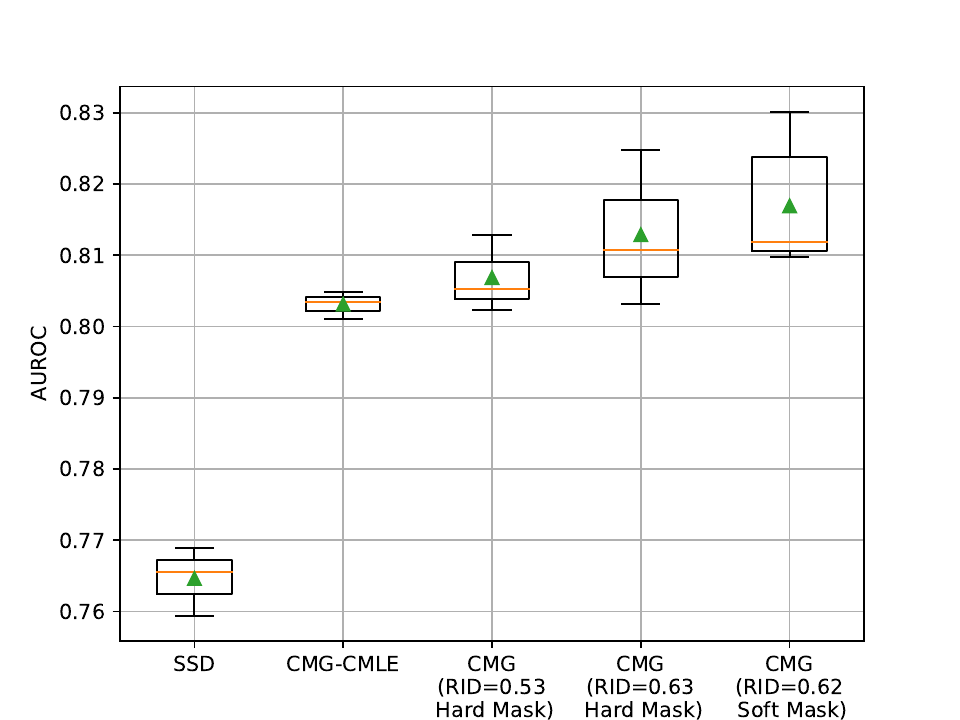}
		\caption{The box plot of CMG with different RID and scale values. The green triangle represents the mean of five runs. The orange line represents the median in the five runs.}
		\label{Figure:box}
	\end{figure}
	
	We further demonstrate the importance of RID by Figure \ref{Figure:box}. Figure \ref{Figure:box} illustrates that the median of RID=0.63 for Hard Mask is higher than that of RID = 0.53 for Hard Mask. That is, with the same mask method, the higher the accuracy of RID, the better the performance of CMG. On the other hand, with similar performance of RID, different mask methods also influence the performance of CMG. With Soft Mask, the minimum and the maximum are significantly higher than those of Hard Mask. This indicates that the soft mask method (i.e., keeping some distribution of raw image pixels by timing a small constant) is better than the hard mask method (i.e., replacing raw pixels with 0).

	\subsection{Convergence Analysis}
	This section shows the convergence of the training models in CMG.
	
\begin{figure}[h]
		\includegraphics[width=0.47\textwidth]{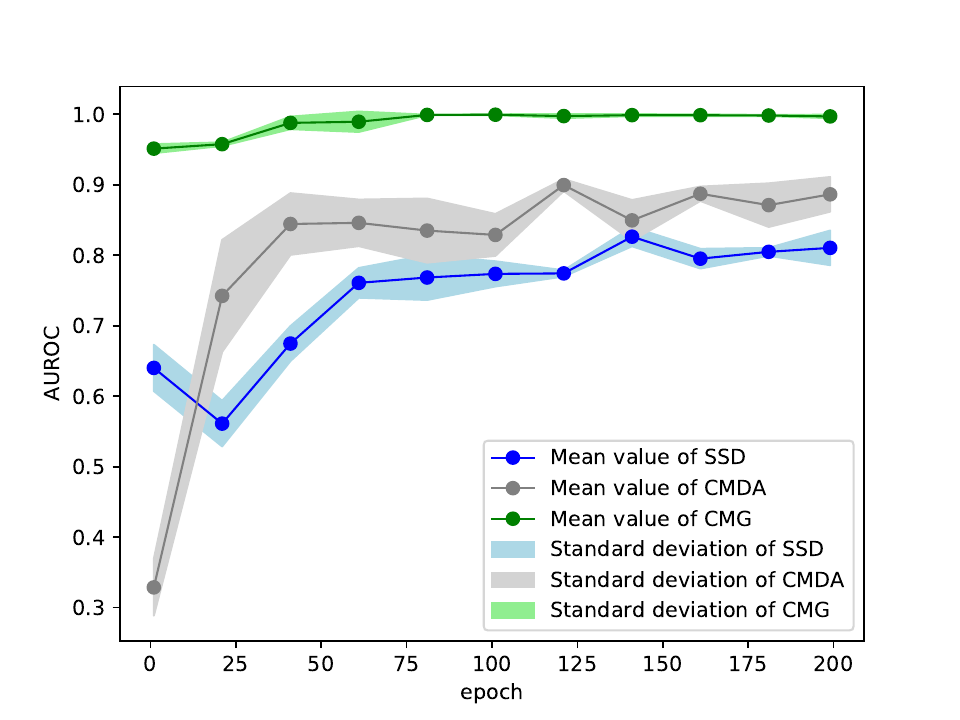}	
		\caption{Convergence of CMG, CMDA and SSD.}	
		\label{Figure:convergence on UCM}
	\end{figure}

	We show the convergence of CMG, CMDA, and SSD in Figure \ref{Figure:convergence on UCM}. Obviously, CMG outperforms CMDA and SSD from scratch. In the first epoch, CMG gets about $95\%$, while the AUROC of other methods is lower than $65\%$.
	Such significant improvement benefits from CMER reducing the redundant information of raw images. With processed images, the model can quickly understand the critical content. Thus, the model achieves excellent performance even after one epoch. On the other hand, after about 80 epochs, the standard deviation of CMG is slight, which indicates that CMG can converge to a stable value compared to CMDA and SSD.
	
	\section{Conclusion}
	In this work, we first analyze the differences between the vision and language modalities using the V-measure and accuracy, based on which we discuss the redundant information and sparse space issues in the vision modality. To address these challenges, we propose the Cross-modal Guidance (CMG), which includes Cross-modal Entropy Reduction (CMER) and Cross-modal Linear Embedding (CMLE). Specifically, with the help of the language modality, CMER masks some useless pixels to make the model focus on critical content. Additionally, CMLE learns a compact latent space for the vision modality with the correlation structure matrix provided by the language modality. We have theoretically and empirically demonstrated the soundness and effectiveness of the proposed method. Our experiments on different datasets, such as Class-COCO, UCM caption, and Wikipedia, show that the proposed method can outperform or achieve highly competitive performance compared to other anomaly detection methods.


	\bibliographystyle{IEEEtran}
	\bibliography{IEEEabrv,sample-base}

	\vfill

\end{document}